\documentclass[11pt,a4paper]{article}
\usepackage[hyperref]{emnlp2018}
\usepackage{times}
\usepackage{latexsym}
\usepackage{caption}
\usepackage{subcaption}
\usepackage{float}
\usepackage{multirow}
\usepackage{graphicx}
\usepackage{amsmath}
\usepackage[many]{tcolorbox}
\usepackage{url}
\usepackage{hhline}
\usepackage{comment}
\usepackage{colortbl}

\DeclareCaptionFont{10pt}{\fontsize{10pt}{12pt}\selectfont}
\newcommand{\rulem}{\textsc{Rule-based}}
\newcommand{\neuralm}{\textsc{Neural}}
\newcommand{\humanm}{\textsc{Human}}
\newcommand*\samethanks[1][\value{footnote}]{\footnotemark[#1]}
\definecolor{light-gray}{gray}{0.9}

\usepackage{url}

\aclfinalcopy 

\title{Transforming Question Answering Datasets \\ Into Natural Language Inference Datasets}

\author{Dorottya Demszky\thanks{\;\;Equal contribution.} \\
  Department of Linguistics \\
 Stanford University\\
  {\tt ddemszky@stanford.edu} \\\And
  Kelvin Guu\samethanks\\
  Department of Statistics \\
  Stanford University \\
  {\tt kguu@stanford.edu} \\\And
  Percy Liang\\
  Department of Computer Science \\
  Stanford University \\
  {\tt pliang@cs.stanford.edu} \\}

\date{}

\begin{document}
\maketitle

\begin{abstract}
Existing datasets for natural language inference (NLI) have propelled research
on language understanding.
We propose a new method for automatically deriving NLI
datasets from the growing abundance of large-scale \emph{question answering} datasets.

Our approach hinges on learning a sentence transformation model
which converts question-answer pairs into their declarative forms.
Despite being primarily trained on a single QA dataset, we show that
it can be successfully applied to a variety of other QA resources.
Using this system, we automatically derive a new freely available dataset of over 500k NLI examples (QA-NLI), and
show that it exhibits a wide range of inference phenomena
rarely seen in previous NLI datasets.


\end{abstract}

\section{Introduction}
\label{sec:intro}



Natural language inference (NLI) is a task that incorporates much
of what is necessary to understand language, such as the ability to leverage world knowledge or perform lexico-syntactic reasoning.
Given two sentences, a premise and a hypothesis, an NLI system must determine whether the hypothesis is implied by the premise.

\begin{figure}[t!]
 \centering
   \centering
   \includegraphics[width=.95\linewidth]{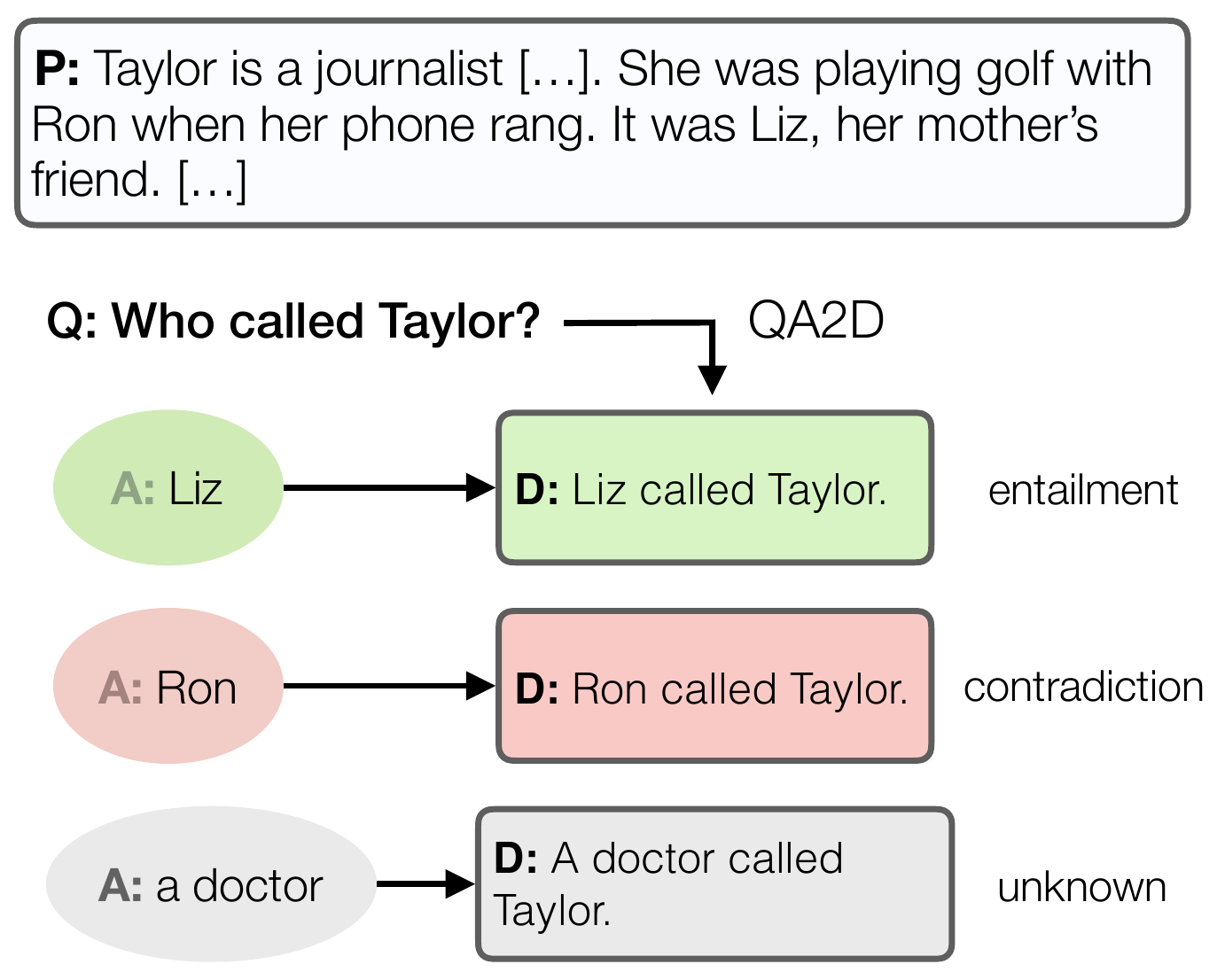}
   \caption{We learn a mapping from a question-answer pair into a declarative sentence (QA2D), which allows us to convert question answering datasets into natural language inference datasets.}
   \label{fig:example}
\end{figure}

\begin{table*}[t!]
\centering
\resizebox{.85\textwidth}{!}{%
\begin{tabular}{|l|l|l|l|l|l|}
\hline
\textbf{Properties}    & \textbf{MovieQA} & \textbf{NewsQA} & \textbf{QAMR}        & \textbf{RACE} & \textbf{SQuAD} \\ \hline
\# wh questions & 11k              & 100k            & 100k                 & 20k           & 100k           \\ \hline
Domain                 & Movie plots      & CNN             & Wikinews + Wikipedia & English exams & Wikipedia      \\ \hline
Multiple choice        & yes              & no              & no                   & yes           & no             \\ \hline
Answer type            & free-form            & span            & span                 & free-form          & span           \\ \hline
Passage type          & 1-3 sentences        & mult. par.      & sentence             & paragraph     & paragraph      \\ \hline
Avg question length    & 10.7             & 7.6             & 6.7                  & 11            & 11.5           \\ \hline
Avg word overlap       & 46\%             & 73\%            & 50\%                 & 62\%          & 75\%           \\ \hline
\end{tabular}%
}
  \caption{Properties of the different QA datasets that we evaluated on. Together they cover a wide range of domains and evidence (passage) types, from sentences to multiple paragraphs and they add up to about 500k NLI examples including the multiple choice QA options. The average question length is counted by tokens and the average word overlap is measured by the percentage of tokens from the question that appear in the evidence.}
\label{tab:qa_props}
\end{table*}

Numerous datasets have emerged to evaluate NLI systems \citep{marelli2014sick, pavlick2016most, lai2017natural}. Two of the largest ones, SNLI \citep{bowman2015large} and MultiNLI \citep{williams2017broad} are rich in various linguistic phenomena relevant to inference (e.g. quantification and negation), but they lack certain other phenomena, such as multi-sentence reasoning, which can be important for various downstream applications. 

In this paper, we propose to augment and diversify NLI datasets by automatically deriving
large-scale NLI datasets from existing \emph{question
answering} (QA) datasets, which have recently become abundant and capture a wide range of reasoning
phenomena.\footnote{Data and code are available here:\\ \url{https://bit.ly/2OMm4vK}} 

Inspired by the connection between QA and NLI noted by \citet{dagan2006pascal},
we take the following approach: given a passage of text, a
question about it ($Q$: \textit{Who called Taylor?}) and an answer ($A$:
\textit{Liz}), we perform sentence transformations to combine the question and
answer into a declarative answer sentence ($D$: \textit{Liz called Taylor}).
We then observe that the passage and the declarative sentence form a (premise,
hypothesis) NLI pair. This is illustrated in Figure~\ref{fig:example}, and
elaborated on in Section~\ref{sec:methods}, where we also discuss how to
generate negative (non-entailed) examples. This approach is similar to the way SciTail \citep{khot2018scitail} was constructed, except that our method is fully automated.

Deriving NLI from QA has two key advantages. First, large-scale QA datasets
are abundant. Second, existing QA datasets cover a wide range of
reasoning strategies, which we can now import into the study of NLI. Both
advantages likely stem from the fact that question answering and question
formulation are organic tasks that people perform in daily life ---
making QA data easy to crowdsource \citep{he2015question}, and easy to find in well-designed
pre-existing resources such as reading comprehension exams \citep{lai2017race}. In
Section~\ref{sec:datasets}, we describe the QA datasets we work with.

Deriving a declarative sentence from a question-answer pair is the key step of our approach, a subtask
that we call QA2D. We explore three different ways to perform QA2D:
(i) a rule-based system (Section~\ref{sec:qa2d_rule}),
(ii) crowdsourcing (Section~\ref{sec:qa2d_crowd}) and (iii) a
neural sequence model (Section~\ref{sec:qa2d_neural}).

These three approaches build on each other:
we demonstrate that a good rule-based system can accelerate and improve the quality of crowdsourcing (by providing workers with an initial draft) while not introducing any systematic bias. This enables us to collect a dataset of 100,000 $(Q, A, D)$ triples,
which we then use to train a neural QA2D model. Although
this model is primarily trained on triples from SQuAD ~\cite{rajpurkar2016squad},
we show that our model generalizes very well to other QA datasets spanning a
variety of domains, such as Wikipedia, newswire and movie plots. Our automatically generated declaratives exactly match the human gold answer 45--57\% of the time, with BLEU scores ranging between 73--83, depending on the dataset (Section~\ref{sec:qa2d_results}).

With our automated QA2D system in place, we apply it to five different QA datasets, creating over 500,000
NLI examples, which we make freely available. Given the diverse nature of the QA datasets we use, the resulting NLI dataset (QA-NLI) also exhibits a wide range of different inference phenomena, such as multi-sentence and meta-level reasoning and presupposition-based inference. We perform a thorough analysis of the
resulting phenomena, quantifying this diversity both
in terms of the \emph{type of reasoning} and the \emph{contextual scope} required
to perform that reasoning. We also conduct other analyses that suggest that our approach can eliminate some of the \emph{annotation artifacts} \citep{gururangan2018annotation} present in SNLI and MultiNLI.

\section{Approach}
\label{sec:methods}

We now formally define our framework for converting a QA example into
an NLI example, including how to generate negative (non-entailed) NLI examples.

A QA example contains a passage of text $P$, a question $Q$ regarding the text and
an answer span $A$, as illustrated in Figure~\ref{fig:example}. We perform
sentence transformations (QA2D) to combine $Q$ and $A$ into a declarative
answer sentence $D$. We then simply recognize that if $A$ is a correct answer,
then $(P, D)$ is an entailed NLI pair.

Alternatively, if $A$ is an incorrect answer or $Q$ cannot be answered
using the information in $P$, then $D$ is \emph{not} implied by $P$,
yielding a negative NLI pair.
Incorrect answers are available in QA datasets featuring multiple choice
answers, such as MovieQA~\citep{tapaswi2016movieqa}, RACE~\citep{lai2017race} and MCTest~\citep{richardson2013mctest}. Unanswerable questions are available in SQuADRUn~\citep{rajpurkar2018squadrun} and we expect the number of such datasets to grow with the
advancement of QA research.

\paragraph{Inference labels.}
In existing NLI datasets, examples are labeled with
one of three relations: entailment, neutral/unknown or contradiction. When
performing automated QA2D, we can only make a two-way distinction between
entailment and non-entailment.\footnote{
 It is tempting to think that incorrect answers yield contradiction labels, while unanswerable questions yield neutral labels. Unfortunately, this is false. Figure~\ref{fig:example} illustrates an example where an incorrect multiple choice answer does not yield a contradiction. As for unanswerable questions, one example would be: $P$: ``The score was negative.'', $Q$: ``What was the exact score?'' (unanswerable), $A$: ``10'' (yields contradiction, not neutral).
 }

\paragraph{Weakly supervised QA datasets.}
In many QA datasets, the passage $P$ is a short paragraph (e.g. SQuAD~\citep{rajpurkar2016squad}) or even
a single sentence (e.g. QAMR~\citep{michael2018qamr}). This yields a short, simple premise in the
resulting NLI example. However, some weakly supervised QA datasets such as
NewsQA~\citep{trischler2017newsqa}, RACE and TriviaQA~\citep{joshi2017triviaqa} choose $P$ be an entire document
or even corpus of documents. In this case, the resulting NLI pair's
premise could be large, but is still valid. In
Table~\ref{tab:qa_props}, we describe the ``passage type'' for each QA dataset
we work with.

\begin{figure}[t]
 \centering
   \centering
   \includegraphics[width=0.95\linewidth]{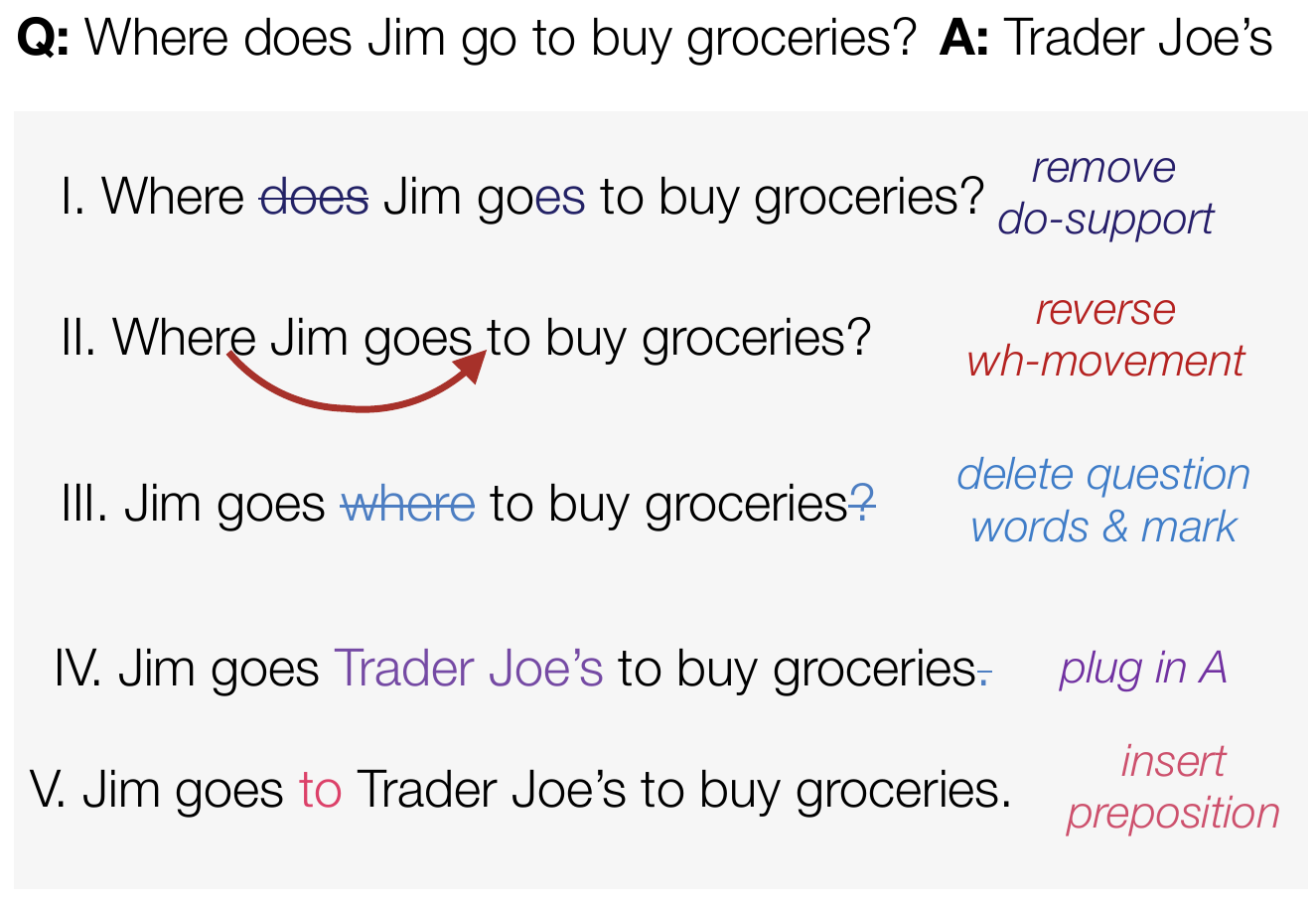}
   \caption{An illustration of the syntactic transformations needed to perform QA2D. In this example, to perform step II one needs to know that \textit{where} is complement of \textit{go} and not that of \textit{buy}, and to perform step V, one needs to chose the appropriate preposition to insert.}
   \label{fig:rulebased}
\end{figure}



\vspace{-.4em}
\section{Datasets}
\label{sec:datasets}

Table~\ref{tab:qa_props}
summarizes the properties of the five QA datasets we transform into NLI: MovieQA,
NewsQA,
QAMR and RACE and SQuAD. When choosing the datasets, we sought to maximize the structural and topical diversity of our data. 

The domains of these datasets include movie plots, newswire text, Wikipedia and English exams that cover a wide range of genres and topics. The \emph{passage types} range from a sentence to multiple paragraphs\footnote{For MovieQA, we only used the plot summaries as the evidence, but one could easily use the full movie scripts or audiovisual data as well.} and the answer type may be either a substring (span) within the passage or free-response text. The questions vary greatly in terms of their type and difficulty: questions in QAMR hinge on selecting the right arguments within a single sentence, while questions in RACE, written for middle- and high-schoolers, require holistic reasoning about the text (e.g. \textit{What is the main message of the passage?}).

\section{QA2D: Rule-based}
\label{sec:qa2d_rule}

At first glance, QA2D appears to be a highly structured task guided by clear rules ---
indeed, the reverse problem of converting declarative sentences into questions
is often taught in grammar textbooks.
However, this belies the many nuanced semantic decisions that are effortlessly made by
native English speakers, yet challenging to codify. For example, non-native speakers find
it notoriously hard to prepend the right prepositions/articles before phrases, as there are
no simple rules.

To demonstrate these challenges, we develop a strong rule-based system (see Section~\ref{sec:qa2d_results} for results) to test how
far we can go towards solving QA2D. The main steps of this system are illustrated in Figure~\ref{fig:rulebased}.

The success of the rule-based model hinges on part-of-speech tagging and parsing accuracy, given that we need to correctly identify the wh-
word, the root word, any auxiliary or copula, as well as prepositions and particles that are dependents of the wh-word or the root. We used the state-of-the-art Stanford Graph-Based Neural Dependency Parser~\citep{dozat2017stanford} to POS tag and parse $Q$ and $A$. We found that about 10\% of the mistakes made by our rule-based system are due to tagging/parsing errors.\footnote{The main errors made by the tagger/parser include tagging a verb as a noun, which is prevalent because in the presence of do-support, the inflections are removed from the main verb (e.g. \emph{When did the war \textbf{end}?}). Another class of parsing errors is identification of the parent of a dangling preposition/particle (e.g. \emph{Which friend did Olga \textbf{send} a letter \textbf{to} last week?}).}

We encountered several semantic idiosyncrasies that proved difficult to account for by rules. For example, if the answer span is a bare named entity (i.e. without an article) referring to an organization/institution, generally it is okay to leave it bare (e.g. \textit{Sam works at WHO.}), but sometimes a definite article needs to be inserted (e.g. \textit{Sam works at the UN}).

\section{QA2D: crowdsourced}
\label{sec:qa2d_crowd}

\begin{table}
\centering
\resizebox{.4\textwidth}{!}{%
\begin{tabular}{|l|l|l|l|l|}
\hline
\textbf{Split}         & \textbf{Source}        & \textbf{\# Ex.} & \textbf{\# Ann.} & \textbf{Setup}                                               \\ \hline
\multirow{2}{*}{train} & SQuAD                  & 68986           & 1                & \begin{tabular}[c]{@{}l@{}}E (74\%)\\ S (26\%)\end{tabular} \\ \cline{2-5} 
                       & Other                  & 4$\times$1000        & 1                & S                                                           \\ \hline
\multirow{3}{*}{dev}   & \multirow{2}{*}{SQuAD} & 7350            & 1                & S                                                            \\ \cline{3-5} 
                       &                        & 1000            & 3                & S                                                            \\ \cline{2-5} 
                       & Other                  & 4$\times$500         & 1                & S                                                            \\ \hline
\multirow{3}{*}{test}  & \multirow{2}{*}{SQuAD} & 7377            & 1                & S                                                            \\ \cline{3-5} 
                       &                        & \cellcolor[HTML]{E7E7E7}     1000      & 3                & S                                                            \\ \cline{3-5}  & Other     & \cellcolor[HTML]{E7E7E7} 4$\times$1000       & 3                & S                                                            \\ \hline
\end{tabular}%
}
  \caption{The composition of our collected data. \# Ex. and \# Ann. refer to the number of unique examples and to the number of annotations (gold answers) per example, respectively. The last column lists the setup that was used for collecting the examples: post-editing (E) or from scratch (S). ``Other'' denotes the four other  QA datasets besides SQuAD. The gray indicates the examples that we used for our evaluations in Section~\ref{sec:qa2d_results}.
  }
\label{tab:coll_data}
\end{table}

\begin{figure*}[t!]
\centering
  \centering
  \includegraphics[width=0.95\linewidth]{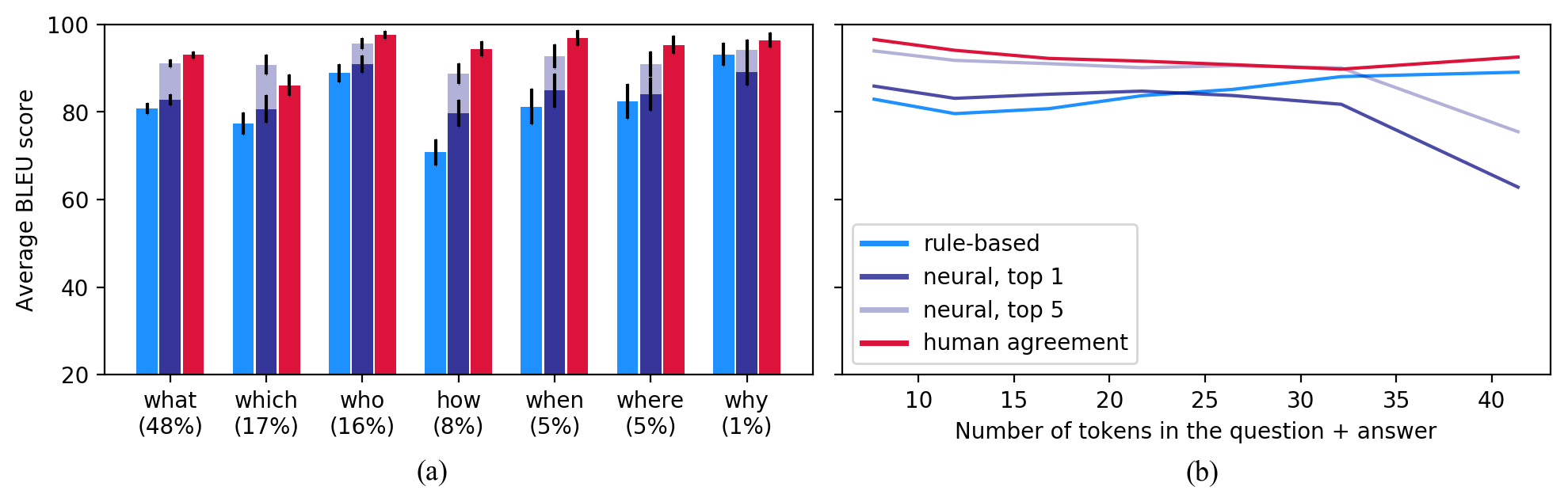}
  \caption{
    Figure (a) shows the results based on question type (brackets indicate their proportion in the data) and Figure (b) shows the results based on the length of $Q + A$. The error bars denote the 99\% confidence interval for the true expected performance of each model (randomness comes from noise in Turker annotations and the random sampling of the evaluation set). Note: human and model scores should not be directly compared. Human agreement is maximum BLEU score when comparing each of 3 human annotations against the two others, while the models' outputs are compared against the 3 human annotations. We include human results to quantify variation across human annotators. }
  \label{fig:data_props}
\end{figure*}

Even though our rule-based model is reasonably strong, it is far from perfect. We decided to build a supervised neural model, which required the collection of human-authored gold declarative sentences. We describe our data collection method (Section~\ref{ssec:crowd_method}) and the distribution of our collected data across the five QA datasets (Section~\ref{ssec:crowd_qa_distr}).

\subsection{Data Collection}
\label{ssec:crowd_method}

We crowdsourced the QA2D task on Amazon Mechanical Turk using two different setups. In Setup S, Turkers were presented with $Q$ and $A$, then asked to write a full sentence answer $D$ from \textbf{\underline{S}}cratch. In Setup E, instead of writing $D$ from scratch, Turkers were asked to \textbf{\underline{E}}dit the output of our rule-based system (see Section~\ref{sec:qa2d_rule}) until it is a well-formed
sentence. Turkers were not provided with the supporting passage $P$ in either setup because we wanted to prevent them from including information in $D$ that is not supported by $Q$.
%

\paragraph{Writing from scratch vs post-editing.} There is a trade-off between the two setups: while Setup S minimizes bias towards the rule-based output, writing from scratch takes more time and leaves room for more typos than post-editing. Indeed, when comparing 100 random examples generated from each setup, we found that 91\% of Setup S outputs were valid (grammatical and complete
%
%
while 97\% of Setup E outputs were valid. However, since Setup E could potentially bias our data, we exclusively used Setup S for collecting all evaluation data.

\subsection{Distribution of Source QA Datasets}
\label{ssec:crowd_qa_distr}
We decided to select one QA dataset among the five QA datasets to collect the majority of our data, so that we could test the ability of our neural model (Section~\ref{sec:qa2d_neural}) to generalize to other datasets. We chose SQuAD to be the main source of QA pairs because of its large size, high quality and syntactic diversity. We limited ourselves to its
training set for the data collection, filtering out non-wh-questions,
which left us
with a total of 85,713 QA pairs.
\footnote{The random split between train (80\%), dev (10\%) and test (10\%) sets was made based on Wikipedia article titles corresponding to the QA pairs.}
In addition, we randomly sampled a smaller set of QA pairs from the four other datasets, most of which were used for evaluation. 

Table~\ref{tab:coll_data} summarizes the composition of our newly collected dataset of gold declarative answer sentences. For each of the dev and test sets, we collected three annotations for 1000 examples to account for the fact that there can be multiple possible correct QA2D transformations. The distribution of datasets within the three data splits are: train (95\% SQuAD, 5\% other four), dev (81\% SQuAD, 19\% other four) and test (20\% for each five datasets).

\section{QA2D: Neural Sequence Model}
\label{sec:qa2d_neural}
In Section~\ref{sec:qa2d_rule}, we discussed some of the issues (mostly
involving semantics) that our rule-based system cannot handle. To
improve over this baseline, we develop a neural sequence generation model to perform
QA2D.

From the crowdsourcing described in the previous section, we have a dataset of
($Q$, $A$, $D$) tuple. We use this to learn a model of
$p(D \mid Q, A)$, implemented with an encoder-decoder architecture.
The inputs $Q$ and $A$ are each encoded using a bidirectional three-layer LSTM \citep{hochreiter1997lstm} (the same
encoder weights are used for both inputs). $D$ is then generated using a
three-layer LSTM decoder equipped with one attention head \citep{bahdanau2015neural}
for each input, and a copy mechanism based on \citet{gu2016copying}.\footnote{The copy mechanism is similar to \citet{gu2016copying}, except that out-of-vocabulary copyable words are represented using absolute positional embeddings, rather than the encoder's hidden state at the position of the copyable word.}
Word embeddings
are initialized with GloVe \citep{pennington2014glove}. The model is then
trained using a standard cross entropy loss minimized with Adam \citep{kingma2014adam}.

\vspace{-.4em}
\section{QA2D: Results}
\label{sec:qa2d_results}
\bgroup
\def\arraystretch{1.2}
\begin{table*}[]
\centering
\resizebox{\textwidth}{!}{%
\begin{tabular}{llll}
\hline
\textbf{Question}                                                                                          & \textbf{Answer}                                                              & \textsc{Rule-based}                                                                                                                            & \textsc{Neural}                                                                                                                           \\ \hline \hline
When was Madonna born?                                                                                     & August 16, 1958                                                              & \cellcolor[HTML]{FFCCC9}Madonna was born in August 16, 1958.                                                                                   & \cellcolor[HTML]{BCF4BC}Madonna was born on August 16, 1958.                                                                              \\ \hline
\begin{tabular}[c]{@{}l@{}}Who asks who to hit them\\ outside of the bar?\end{tabular}                     & \begin{tabular}[c]{@{}l@{}}Tyler asks the\\ Narrator to hit him\end{tabular} & \cellcolor[HTML]{FFCCC9}\begin{tabular}[c]{@{}l@{}}Tyler asks the Narrator to hit him asks\\ who to hit them outside of the bar.\end{tabular} & \cellcolor[HTML]{BCF4BC}\begin{tabular}[c]{@{}l@{}}Tyler asks the narrator to hit them outside\\ of the bar.\end{tabular}                 \\ \hline
\begin{tabular}[c]{@{}l@{}}What surprising fact do the guys\\ learn about Jones?\end{tabular}              & \begin{tabular}[c]{@{}l@{}}That he has never\\ killed anyone\end{tabular}    & \cellcolor[HTML]{FFCCC9}\begin{tabular}[c]{@{}l@{}}The guys learn that he has never\\ killed anyone about Jones.\end{tabular}                 & \cellcolor[HTML]{BCF4BC}\begin{tabular}[c]{@{}l@{}}The guys learned about Jones that he has\\ never killed anyone.\end{tabular}           \\ \hline
Where is someone overlooked?                                                                               & American society                                                             & \cellcolor[HTML]{BCF4BC}\begin{tabular}[c]{@{}l@{}}Someone is overlooked in American\\ society.\end{tabular}                                   & \cellcolor[HTML]{FFCCC9}\begin{tabular}[c]{@{}l@{}}Someone overlooked is in American\\ society.\end{tabular}                              \\ \hline
\begin{tabular}[c]{@{}l@{}}When did Johnson crash\\ into the wall?\end{tabular}                            & \begin{tabular}[c]{@{}l@{}}halfway through\\ the race\end{tabular}           & \cellcolor[HTML]{BCF4BC}\begin{tabular}[c]{@{}l@{}}Johnson crashed into the wall \\ halfway through the race.\end{tabular}                   & \cellcolor[HTML]{FFCCC9}\begin{tabular}[c]{@{}l@{}}Johnson shot into the wall halfway \\ through the race.\end{tabular}                  \\ \hline
\begin{tabular}[c]{@{}l@{}}What is an example of a corporate \\ sponsor of a basketball team?\end{tabular} & Marathon Oil                                                                 & \cellcolor[HTML]{BCF4BC}\begin{tabular}[c]{@{}l@{}}An example of a corporate sponsor of \\ a basketball team is Marathon Oil.\end{tabular}     & \cellcolor[HTML]{BCF4BC}\begin{tabular}[c]{@{}l@{}}Marathon Oil is an example of a corporate\\ sponsor of a basketball team.\end{tabular} \\ \hline
Where was the baby found?                                                                                  & \begin{tabular}[c]{@{}l@{}}onboard a Carnival\\ cruise ship,\end{tabular}    & \cellcolor[HTML]{FFCCC9}\begin{tabular}[c]{@{}l@{}}The baby was found in onboard a\\ Carnival cruise ship.\end{tabular}                        & \cellcolor[HTML]{FFCCC9}\begin{tabular}[c]{@{}l@{}}The baby was found at onboard a Carnival\\ cruise ship.\end{tabular}                   \\ \hline
\end{tabular}%
}
\caption{A randomly picked sample of those outputs where $\neuralm \neq \rulem$. Green indicates good output(s) and red indicates bad ones.}
\label{tab:sample_outputs}
\end{table*}
\egroup

In this section, we assess the performance of our rule-based ($\rulem$) and neural ($\neuralm$) QA2D systems, using both automated metrics and human evaluation.
The evaluations are conducted on the test set formed from all five QA datasets (gray cells in
Table~\ref{tab:coll_data}) where each example includes three human
annotations. 

\begin{table*}[ht]
\begin{minipage}[b]{0.62\linewidth}
\centering
\resizebox{\textwidth}{!}{%
\begin{tabular}{|c|c|c|c|c|c|c|c|c|c|c|c|}
\hline
\multicolumn{2}{|c|}{Dataset}  & \multicolumn{2}{c|}{MovieQA} & \multicolumn{2}{c|}{NewsQA}        & \multicolumn{2}{c|}{QAMR}          & \multicolumn{2}{c|}{RACE} & \multicolumn{2}{c|}{SQuAD}         \\ \hline
\multicolumn{2}{|c|}{Model}    & RM                     & NM  & RM                   & NM          & RM                   & NM          & RM                   & NM & RM                   & NM          \\ \hline
\multirow{2}{*}{Top 1} & BLEU  & 81                     & 84  & 82                   & 83          & 82                   & 85          & 80                   & 83 & 82                   & \textbf{86} \\ \cline{2-12} 
                       & Match & 38                     & 44  & 49                   & 49          & 54                   & \textbf{57}          & 30                   & 44 & 45                   & 53          \\ \hline
\multirow{2}{*}{Top 5} & BLEU  & \multirow{2}{*}{N/A}   & 89  & \multirow{2}{*}{N/A} & 88          & \multirow{2}{*}{N/A} & 90          & \multirow{2}{*}{N/A} & 89 & \multirow{2}{*}{N/A} & \textbf{91} \\ \cline{2-2} \cline{4-4} \cline{6-6} \cline{8-8} \cline{10-10} \cline{12-12} 
                       & Match &                        & 52  &                      & 58 &                      & \textbf{67} &                      & 55 &                      & 62          \\ \hline
\end{tabular}%
}
\caption{BLEU and exact match scores (out of a 100) when comparing the outputs of $\rulem$ (RM) and $\neuralm$ (NM) against the human gold.  }
\label{tab:results}
\end{minipage}\hfill
\begin{minipage}[b]{0.35\linewidth}
\centering
\includegraphics[width=\linewidth]{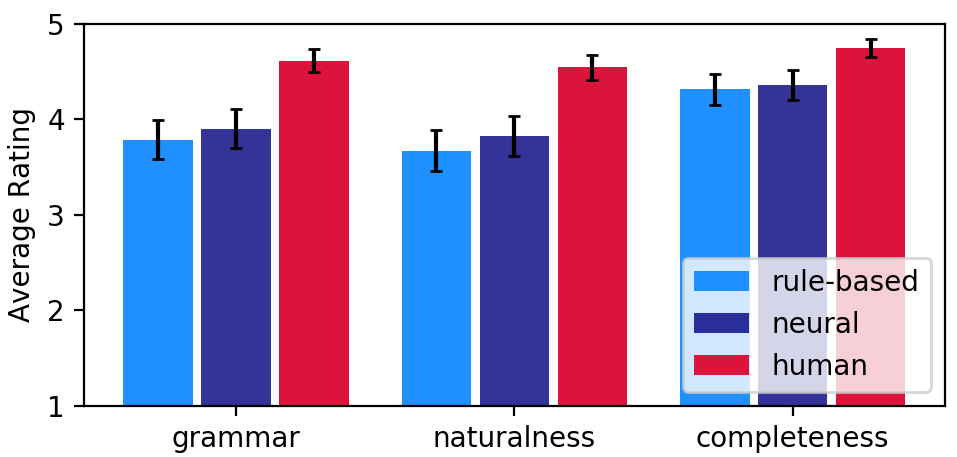}
\captionof{figure}{Human evaluation of the the human, $\rulem$ and $\neuralm$ outputs.}
  \label{fig:human_eval}
\end{minipage}
\end{table*}

\subsection{Quantitative Results}
For our quantitative evaluation, we employed two metrics: BLEU and string match (ignoring case and punctuation). We evaluated $\neuralm$ both on the top 1 output and on the top 5 outputs (max over the beam).

\paragraph{Rule-based vs neural.}  

Overall, the performance of $\neuralm$ is consistently stronger than $\rulem$. From Table~\ref{tab:results}, we see that across datasets, $\neuralm$ leads $\rulem$ by an average of 2.6 BLEU points, and by 6.2\% on exact match accuracy.
$\neuralm$ is also capable of producing a top-5 beam of outputs, and when we evaluate only the best of these 5 outputs,
we observe an almost 30\% improvement in scores. 

We find that predictions of $\neuralm$ and $\rulem$ exactly match 40\% of the time. Table~\ref{tab:sample_outputs} includes examples when the two models' outputs do not match. As we hypothesized, the $\neuralm$ learned semantic patterns, such as preposition/article choice and the removal of redundant words from the answer span, that $\rulem$ was not able to handle.


\begin{table*}[th]
\centering
\resizebox{\textwidth}{!}{%
\begin{tabular}{|llll|}
\hline
\textbf{Reasoning} & \textbf{Description}                                                        & \textbf{Example}                                                                                  & \textbf{Source} \\ \hline \hline
Argument & \begin{tabular}[c]{@{}l@{}}Affects only a single \\argument of a predicate in $T$.\end{tabular}& \begin{tabular}[c]{@{}l@{}}\textit{T: \textbf{Caitlin de Wit}: I ran a little bit, and I rode horses.}\\
\textit{H: Caitlin is de Wit's first name.}\end{tabular}& QAMR \\ \hline
Sentence & \begin{tabular}[c]{@{}l@{}}Affects one or more\\ predicates within a single\\ sentence in $T$.\end{tabular}& \begin{tabular}[c]{@{}l@{}}\textit{T: Michael Jackson will perform 10 concerts in London} \\\textit{in July in what he described Thursday as a "final curtain call." {[}...{]}}\\
\textit{H: Michael Jackson has announced 10 concerts.}\end{tabular}& NewsQA\\ \hline
\begin{tabular}[c]{@{}l@{}}Multi-\\sentence\end{tabular} &\begin{tabular}[c]{@{}l@{}}Affects multiple sentences\\ in $T$.\end{tabular} &\textit{\begin{tabular}[c]{@{}l@{}} T: {[}...{]} invented by the British scientist William Sturgeon {[}...{]}\\ Following Sturgeon's work, {[}...{]} motor {[}...{]} built by {[}...{]} Thomas Davenport {[}...{]} \\The motors ran at up to 600 revolutions per minute {[}...{]}\\
H: Sturgeon and Davenport s motors ran at 600 revolutions per minute .\end{tabular}} & SQuAD \\ \hline \hline
Quantities       & \begin{tabular}[c]{@{}l@{}}Counting or performing other \\ numerical operations; \\ understanding relations \\ between quantities.\end{tabular}     & \textit{\begin{tabular}[c]{@{}l@{}}T: It provided \textbf{more than \$20 billion} in direct financial.\\ H: It only yielded \textbf{\$100,000} in direct financial.\end{tabular}}   & MultiNLI    \\ \hline
 \begin{tabular}[c]{@{}l@{}}Naive\\ physics\end{tabular}       & \begin{tabular}[c]{@{}l@{}}Spatiotemporal/physical \\reasoning  that requires \\a mental simulation of the \\event beyond understanding \\the meaning of the words.\end{tabular}                                          & \textit{\begin{tabular}[c]{@{}l@{}}T: David J. Lavau, 67, of Lake Hughes, California, was found in \\ a ravine a week after losing control of his car on a rural road and \\ \textbf{plunging} 500 feet down an embankment into heavy brush {[}...{]}\\ H: Lavau's car \textbf{came to a rest} 500 feet down an embankment.\end{tabular}}         & NewsQA                  \\ \hline
Attributes         & \begin{tabular}[c]{@{}l@{}}Reasoning about attributes \\ and affordances of entities.\end{tabular}             & \textit{\begin{tabular}[c]{@{}l@{}}T: The situation in Switzerland {[}...{]}. The Swiss German dialects \\ are \textbf{the default everyday} language in virtually every situation {[}...{]}\\ H: Swiss German is the dialect \textbf{spoken} in Switzerland.\end{tabular}}    & SQuAD     \\ \hline
Psychology & \begin{tabular}[c]{@{}l@{}}Making inferences involving\\  people's mental states and \\ attitudes and the way they \\ express them.\end{tabular}  & \textit{\begin{tabular}[c]{@{}l@{}}T: Dear Jorge, {[}...{]} My family are now in Sacramento, California. {[}...{]} \\ \textbf{Before I knew it}, there was hot water shooting up about 60 feet into\\  the air. {[}...{]} \textbf{I'd love to} learn more about this geyser and other geysers {[}...{]}\\ Your friend,Bennetto\\ H: Bennetto's letter expressed \textbf{excitement}.\end{tabular}} & RACE            \\ \hline
Meta               & \begin{tabular}[c]{@{}l@{}}Reasoning about the genre, \\ text structure and author.\end{tabular}              & \textit{\begin{tabular}[c]{@{}l@{}}T: A man and his young son struggle to survive after an \\ unspecified cataclysm has killed most plant and animal life. {[}...{]}\\ H: We meet a man and his young son \textbf{at the beginning of the film}.\end{tabular}}   & MovieQA    \\ \hline
Other       & \begin{tabular}[c]{@{}l@{}}Incorporating any other \\world knowledge.\end{tabular} & \textit{\begin{tabular}[c]{@{}l@{}} T: When asked a \textbf{life lesson} he had to learn the hard way, \textbf{the billionaire said}\\ staying up too late is a habit he is still trying to break. \textbf{``Don't stay up too} \\ \textbf{late} [...] \\ H: Bill Gates gave the \textbf{advice} to avoid staying up too late. \end{tabular}}
    & RACE    \\ \hline
\end{tabular}%
}
\caption{Examples illustrating the different types of reasoning required to determine whether the text $T$ entails the hypothesis $H$.}
\label{tab:phenomena_examples}
\end{table*}

\begin{table*}[ht]
\begin{minipage}[b]{0.62\linewidth}
\centering
\resizebox{\textwidth}{!}{%
\begin{tabular}{cl|r|r|r|r|r||r|}
\cline{3-8}
\multicolumn{1}{l}{}                         &                   & MovieQA & NewsQA & QAMR & RACE & SQuAD & MultiNLI \\ \hline
\multicolumn{1}{|c|}{\multirow{3}{*}{\rotatebox[origin=c]{90}{\centering scope}}} & argument              & 19      &      13  & \textbf{62}   & 1    & 15    & 26       \\ \cline{2-8} 
\multicolumn{1}{|c|}{}                       & sentence          & 56      &   53   & 38   & 14   & 58    & \textbf{72}       \\ \cline{2-8} 
\multicolumn{1}{|c|}{}                       & multi-sentence    & 25      &  34  & 0    & \textbf{85}   & 27    & 2        \\ \hline
\multicolumn{1}{|c|}{\multirow{6}{*}{\rotatebox[origin=c]{90}{\centering type of reasoning}}}  & quantities        & 1       &    1   & 1    & 1    & 2     & \textbf{4}        \\ \cline{2-8} 
\multicolumn{1}{|c|}{}                       & naive physics     & 4       &   \textbf{7}  & 0    & 5    & 4     & 3        \\ \cline{2-8} 
\multicolumn{1}{|c|}{}                       & psych             & 9       &  0    & 1    & \textbf{48}   & 1     & 5        \\ \cline{2-8} 
\multicolumn{1}{|c|}{}                       & meta              & 2       &   5   & 0    & \textbf{12}   & 0     & 1        \\ \cline{2-8} 
\multicolumn{1}{|c|}{}                       & attributes        & 41      &   41 & 12   & \textbf{58}   & 40    & 29       \\ \cline{2-8} 
\multicolumn{1}{|c|}{}                       & other world kn.   & 34      &   15  & 8    & \textbf{68}   & 14    & 5        \\ \hline
\end{tabular}%
}
\caption{Counts for different scope and reasoning types in the five converted QA datasets in comparison to MultiNLI. We manually annotated 100 examples per dataset. While the scope categories are mutually exclusive, the reasoning types are not. }
\label{tab:phenomena_counts}
\end{minipage}\hfill
\begin{minipage}[b]{0.36\linewidth}
\centering
\includegraphics[width=\linewidth]{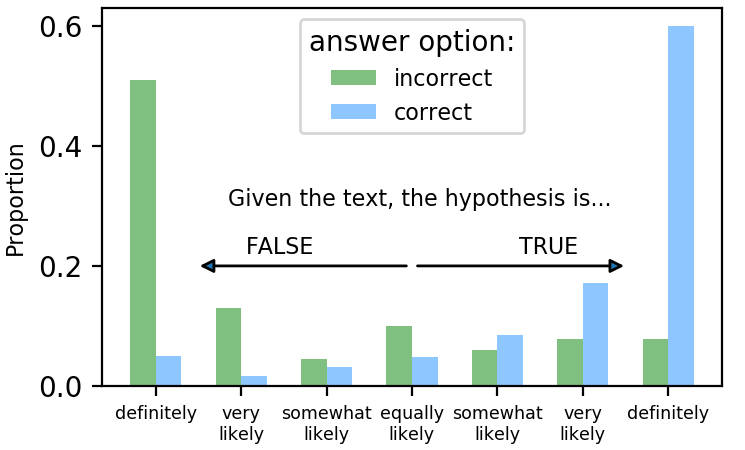}
\captionof{figure}{The distribution of inference ratings for NLI examples based on \textit{incorrect} multiple choice option or \textit{correct} multiple choice option. }
  \label{fig:entail_labels}
\end{minipage}
\end{table*}

\paragraph{Results by dataset.}
Table~\ref{tab:results} breaks down the models' performance by dataset. The first thing to note is the domain-generality of the models --- although BLEU scores are highest on SQuAD, they are only 1-3 points lower on other datasets. As for the exact match, we actually find that both models perform better on QAMR than on SQuAD. This discrepancy is likely due to the length penalty in BLEU, which is affected by shorter answer lengths in QAMR. 

$\rulem$ performs worst on RACE, MovieQA and SQuAD (6-9 points lower on exact match) in comparison to $\neuralm$ due to the fact that the answers in these datasets often require semantically motivated modifications that $\rulem$ cannot handle.  

\vspace{-.4em}
\paragraph{Results by question length.} Figure~\ref{fig:data_props} (b) shows the correlation between the combined length of the question-plus-answer and each model's performance. $\rulem$'s performance is more robust to length increase beyond 30 tokens than $\neuralm$, even if we consider $\neuralm$'s top-5 performance. In contrast, $\neuralm$ does better on inputs shorter than 20 tokens  --- which constitute the majority of the examples --- than on longer ones, which is possibly due to the general tendency of $\neuralm$ to output shorter sequences.

\vspace{-.4em}
\paragraph{Results by question type.} 
In Figure~\ref{fig:data_props} (a) we present the results broken down by question category, which we determined based on the type of wh word in the question. We can see that our models perform best on \textit{who} questions, which is not surprising because there is usually no wh-movement involved when transforming such questions into declaratives (i.e. the wh word simply needs to be replaced). In contrast, the overall performance of the models is the worst on \textit{which} questions, which is most likely due to the fact that the majority of such questions require a decision about which words from the wh phrase to include in $D$ and in what position.  

The only question type for which we see a significant difference between performance of $\rulem$ and that of the neural one is \textit{how} questions (note that \textit{how many}, for example, is considered to be a \textit{how} question). This is because $\rulem$ --- in contrast to $\neuralm$ --- does not copy words from the wh phrase into $D$, which mainly affects \textit{how} questions negatively, given that \textit{how many} tends to be followed by at least one noun (e.g. \textit{how many people}). 
\vspace{-.4em}
\subsection{Human Evaluation} We crowdsourced the evaluation of $\rulem$ and $\neuralm$ on a sample of 100 QA examples for each of the five datasets. For reference, we also treated human outputs as a third system ($\humanm$). In each Turk task, 
a Turker is presented with three outputs (one from each system, in a randomly shuffled order) and is asked to rate them (given a question and an answer span) with respect to three criteria: grammaticality, naturalness,
and completeness.\footnote{Description of ratings: \\
grammaticality: \textit{1 -- Extremely poor, 2 -- Poor, 3 -- OK but has some issue(s), 4 -- Good but slightly unnatural, 5 -- Good} \\
naturalness: \textit{1 -- Extremely unnatural, 2 -- Unnatural, 3 -- OK in some contexts, 4 -- Natural, but could be more so, 5 -- Very natural} \\completeness:  \textit{1 -- Lacks many important words from the question or the answer, 2 -- Lacks a few important words from the question or the answer, 3 -- The sentence is missing one or two words that would add more information, but they aren't necessary, 4 -- The sentence is missing one or two words but it still conveys the same meaning without them, 5 -- The sentence is maximally complete in terms of words (regardless of grammaticality)}

} The results are shown in Figure~\ref{fig:human_eval}.

For grammar, the models' outputs are lower than $\humanm$ (4.6), at 3.8 for $\rulem$ and 3.9 for $\neuralm$. A score of 4 represents the rating \textit{Good but slightly unnatural}. The same pertains to the naturalness scores as well. In terms of completeness, the rule-based and $\neuralm$ both do relatively well (.2-.3 points lower than the human score of 4.8), which is well above the threshold for retaining the correct meaning, given that a rating of 4 requires there to be no semantic consequences of incompleteness.



\vspace{-.4em}
\section{Analysis of NLI Datasets}
\label{sec:analysis}
In this section, we analyze the various phenomena of the NLI datasets we generated
(Section~\ref{ssec:inference_phenomena}), validate our assumptions
about how an answer's correct/incorrect status determines the resulting inference label (Section~\ref{ssec:inference_labels}) and compare our datasets to others in terms of annotation artifacts (Section~\ref{ssec:artifacts}).

\subsection{Inference Phenomena}
\label{ssec:inference_phenomena}
We manually annotated 100 examples for the \emph{scope} and \emph{type of reasoning} required to make a correct inference classification. The categories and their descriptions, paired with examples are illustrated in Table~\ref{tab:phenomena_examples}. 
Looking at the counts in Table~\ref{tab:phenomena_counts}, we can notice that MovieQA, NewsQA and SQuAD are similar. The majority of examples in these datasets require sentence-level reasoning, and the rest are split between multi-sentence and argument-level reasoning at a roughly 5:3 ratio.  While the counts for the types of reasoning are also very close among these datasets, the slight differences can be explained based on genre. For example, plot summaries often focus on human motivations/emotions (\textit{psych}). MultiNLI is closest to these datasets in terms of phenomena counts as well, except that it involves hardly any multi-sentence reasoning and less world knowledge than the other three.

QAMR is unique in that, by construction, two thirds of the examples only involve argument-level reasoning and none of them involve multi-sentence reasoning. Given that inference pairs in QAMR often involve turning a noun phrase into a predicate, this dataset provides us with a lot of inference pairs that stem from presuppositions --- i.e. entailments that still hold even if the premise is negated (e.g. \emph{Taylor Brown was playing golf outside.} and \emph{Taylor Brown was \textbf{not} playing golf outside.} both presuppose \emph{Taylor's last name is Brown.}). 

RACE, in sharp contrast to QAMR, mostly includes entailments that require multi-sentence reasoning. In addition, inference pairs in RACE make extensive use of world knowledge, meta-level reasoning (e.g., about the genre, author's intentions, etc.), and reasoning about human psychology (e.g., a character's reaction to an event).

\subsection{Inference Labels}
\label{ssec:inference_labels}

In Section~\ref{sec:methods}, we made the assumption that inference pairs generated from incorrect answers will be non-entailments. To verify this hypothesis, we crowdsource the inference labeling of 2000 inference pairs based on MovieQA, half of which are generated from correct answer options and half from incorrect ones.
\footnote{For the 1000 examples based on correct answers, we used the test set we already collected. We collected the 1000 examples for incorrect answers using the same QA pairs as for the correct answers. The incorrect answer, in the case of each example, was randomly chosen among the incorrect options.}
In the task, each Turker was provided with five randomly selected premise-hypothesis pairs $(T, H)$ and was asked to rate how likely $H$ is true given $T$.

Figure~\ref{fig:entail_labels} shows the distribution of inference ratings with the examples separated based on incorrect/correct answer options as their source. The human ratings show a strong correlation between answer type and inference score --- as for correct ones, more than 90\% of the examples are more likely to be true than false and as for incorrect ones, about 80\% are not likely to be true.  These findings also support a non-binary notion of entailment, as we can see that about half of the examples in both categories  do not fit into the strict entailment/contradiction dichotomy.

\subsection{Annotation Artifacts}
\label{ssec:artifacts}
\begin{figure}[]
\includegraphics[width=.8\linewidth]{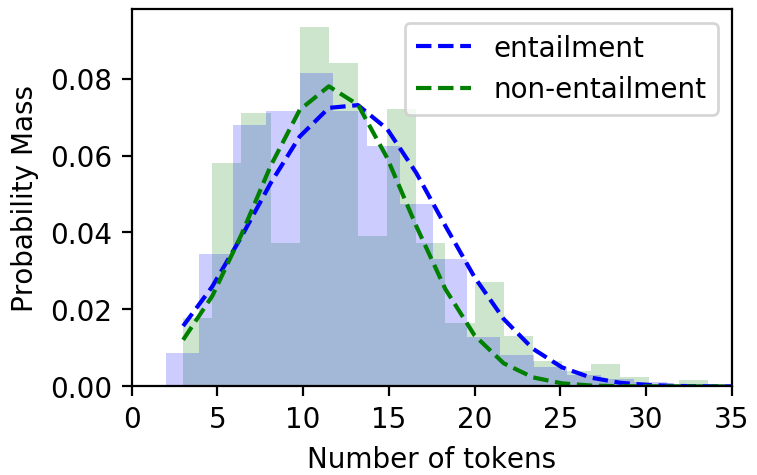}
   \caption{The distribution of the length of NLI examples, generated from MovieQA. Unlike in SNLI and MultiNLI, we found little correlation between the inference label and sentence lengths.}
   \label{fig:entail_lengths}
\end{figure}
\begin{table}[]
\centering
\resizebox{.4\textwidth}{!}{%
\begin{tabular}{|c|c|c|c|}
\hline
\textbf{}             & \textbf{entailment} & \textbf{neutral} & \textbf{contradiction} \\ \hline 
\multirow{5}{*}{MQA}                   & find                & \multicolumn{2}{c|}{all}                   \\
                      & take                & \multicolumn{2}{c|}{police}                \\
                      & years               & \multicolumn{2}{c|}{york}                  \\
                      & son                 & \multicolumn{2}{c|}{car}                   \\
                      & shoot               & \multicolumn{2}{c|}{can}                   \\ \hline
\multirow{5}{*}{SNLI} & outdoors            & tall             & nobody                 \\
                      & least               & first            & sleeping               \\
                      & instrument          & competition      & no                     \\
                      & outside             & sad              & tv                     \\
                      & animal              & favorite         & cat                    \\ \hline
  \multirow{5}{*}{MNLI} & some           & also             & never            \\  & yes                 & because          & no                     \\
                      & something           & popular          & nothing                \\
                      & sometimes           & many             & any                    \\
                      & various             & most             & none \\ \hline                
\end{tabular}%
}
\caption{Top 5 words ordered based on their PMI (\textit{word, class}) with the percentage of examples they occur in in a given \textit{class}, based on the entailments generated from MovieQA (MQA). The statistics for SNLI and MultiNLI (MNLI) are copied from \citet{gururangan2018annotation}.}
\label{tab:pmi_words}
\end{table}

We replicate some of the statistical analyses that ~\citet{gururangan2018annotation} performed on SNLI and MultiNLI to see whether our datasets contain artifacts similar to SNLI and MultiNLI. We perform the analyses on the same dataset we used in Section~\ref{ssec:inference_labels}, based on MovieQA. We ranked the words with the highest PMI(\textit{word, class}) values and found that in our case the non-entailments in our dataset no longer have negation words and the entailments no longer have positive or non-specific words such as those found in SNLI and MultiNLI (Table~\ref{tab:pmi_words}). We also looked at the distribution of hypotheses lengths, separated by label (Figure~\ref{fig:entail_lengths}) and found little/no correlation between the length of an example and its label.

\section{Related Work \& Discussion}

\paragraph{NLI datasets.}
NLI has long served as a testbed for natural language understanding
\citep{dagan2006pascal}. More recently, the emergence of larger-scale datasets
\citep{bowman2015large, williams2017broad}
have also enabled researchers to leverage NLI resources as a rich
source of training data to achieve transfer learning gains on other tasks \citep{conneau2017supervised}. Our datasets are complementary to previous resources, inheriting a rich set of phenomena found in many QA datasets (e.g. high-level reasoning about texts).



\paragraph{QA to NLI.}
Although \citet{white2017inference} and \citet{poliak2018towards} have explored recasting datasets created for various semantic classification tasks (e.g. semantic role labeling and named entity recognition) into NLI datasets, we are the first to perform such an automated conversion on QA datasets.  However,
we are not the first to observe the connection between QA and NLI. In fact,
the seminal work of \citet{dagan2006pascal} employed this connection to
construct a portion of their dataset and so have the creators of SciTail \citep{khot2018scitail}, but performed the QA2D step with human
experts, rather than an automated system. 

\paragraph{Text transformation tasks.}
By ``reformatting'' QA to NLI, we obtain a more generic representation of
inferences: declarative sentences are transformed into other declarative
sentences. This is the same type signature as for sentence simplification
\citep{chandrasekar1996motivations}, paraphrase \citep{lin2001discovery,
bannard2005paraphrasing} and summarization \citep{jones1993might}, highlighting the
close connection between these tasks. Importantly, declarative sentences are
closed under this set of operations, allowing them to be chained together to
perform more complex inferences \citep{kolesnyk2016generating}.

Another related task is question generation \citep{rus2009question}, which could be considered the
reverse of QA2D, although the focus is on selecting interesting questions,
rather than robust sentence transformation.

\paragraph{Neural sequence generation.}
Our QA2D system could be implemented by any general-purpose sequence
generation model. With rapid progress on better generation architectures
\citep{gehring2017convolutional, vaswani2017attention}, we believe it should be possible to further
increase the data efficiency and performance, especially by
leveraging models that incorporate syntactic structure \citep{chen2017improved, eriguchi2017learning} or
a more transducer-like structure \citep{graves2012sequence, yu2016online}.

\paragraph{Future systems.}
Finally, we hope that by increasing the scale of NLI training resources, we
can enable the development of a large variety of new systems such as
generative NLI models which can take a premise and generate relevant
hypotheses \citep{kolesnyk2016generating}, sentence decomposition models which can break a sentence into
multiple entailed parts, and sentence synthesis models which can stitch
multiple pieces back into an entailed whole.

\section*{Acknowledgments}
We would like to thank Chris Potts and members of the Stanford NLP group for their valuable feedback. 

\bibliography{all}
\bibliographystyle{acl_natbib_nourl}

\end{document}